\documentclass{article}
\usepackage{PRIMEarxiv}
\usepackage{amsmath}
\usepackage[utf8]{inputenc} 
\usepackage[T1]{fontenc}    
\usepackage{hyperref}       
\usepackage{url}            
\usepackage{booktabs}       
\usepackage{amsfonts}       
\usepackage{nicefrac}       
\usepackage{microtype}      
\usepackage{lipsum}
\usepackage{fancyhdr}       
\usepackage{graphicx}       
\graphicspath{{media/}}     
\usepackage{subcaption}
\title{Transfer learning for improved generalizability in causal physics-informed neural networks for beam simulations 
\thanks{\emph{Preprint submitted to Elsevier}} 
}

\author{
  Taniya Kapoor, Hongrui Wang, Alfredo N\'u\~nez, Rolf Dollevoet \\
  Department of Engineering Structures\\ Faculty of Civil Engineering and Geosciences\\ Delft University of Technology\\ 
    The Netherlands\\
  \texttt{\{t.kapoor, h.wang-8, a.a.nunezvicencio, r.p.b.j.dollevoet\}@tudelft.nl} \\
}

\begin{document}
\maketitle

\begin{abstract}
This paper introduces a novel methodology for simulating the dynamics of beams on elastic foundations. Specifically, Euler-Bernoulli and Timoshenko beam models on the Winkler foundation are simulated using a transfer learning approach within a causality-respecting physics-informed neural network (PINN) framework. Conventional PINNs encounter challenges in handling large space-time domains, even for problems with closed-form analytical solutions. A causality-respecting PINN loss function is employed to overcome this limitation, effectively capturing the underlying physics. However, it is observed that the causality-respecting PINN lacks generalizability. We propose using solutions to similar problems instead of training from scratch by employing transfer learning while adhering to causality to accelerate convergence and ensure accurate results across diverse scenarios. Numerical experiments on the Euler-Bernoulli beam highlight the efficacy of the proposed approach for various initial conditions, including those with noise in the initial data. Furthermore, the potential of the proposed method is demonstrated for the Timoshenko beam in an extended spatial and temporal domain. Several comparisons suggest that the proposed method accurately captures the inherent dynamics, outperforming the state-of-the-art physics-informed methods under standard $L^2$-norm metric and accelerating convergence.
\end{abstract}


\section{Introduction}
\begin{figure}[h]
\centerline{\includegraphics[width=0.5\columnwidth]{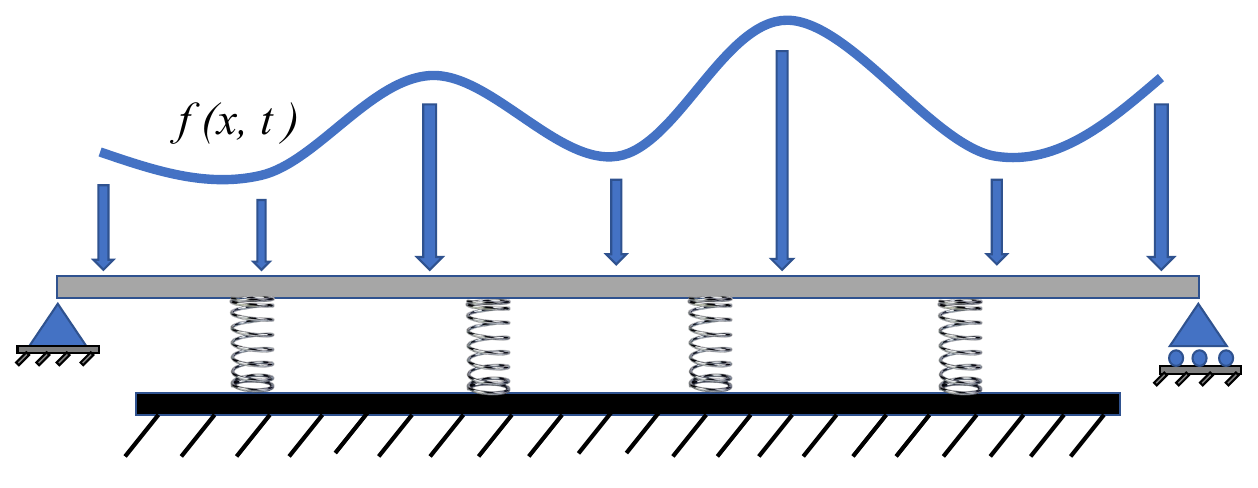}}
\caption{Simply supported beam on an elastic foundation under varying transverse force}
\label{fig1}
\end{figure}

Beams on elastic foundations (Fig.~\ref{fig1}) are a fundamental and indispensable structural component in civil engineering, providing critical support and stability to different and diverse structures (\cite{lamprea2022beams, deng2023dynamic, tsudik2012analysis,hetenyi1946beams}). Due to their characteristic to distribute loads, mitigate deformations, and enhance structural stability, these beams are extensively utilized in various structures, such as railway tracks (\cite{lamprea2022beams}), pile foundations embedded in soils (\cite{petrosian2022analysis}), and longitudinal fibers in a composite elastomer (\cite{tsudik2012analysis}), among others. Understanding their dynamics is essential for ensuring the structural integrity of these systems, developing effective maintenance strategies, optimizing machine performance, refining design methodologies, and enabling precise control mechanisms. These issues highlight the need for advanced methodologies to simulate and predict the underlying dynamics of beams on elastic foundations, facilitating safer, more efficient, and reliable structures and systems.

However, accurately predicting the dynamics of beams on elastic foundations through experiments and measurements could be infeasible (\cite{chang1999robust}). Conducting many experiments with varying materials, conditions, and prototypes becomes impractical and prohibitively costly. In practice, finite element-based software provides a viable alternative for simulating such scenarios (\cite{madenci2015finite}). However, these software solutions are restricted in generalization. For instance, even a slight change in the problem domain requires performing the entire new simulation from scratch, including mesh creation and adjustments (\cite{karniadakis2021physics}). This non-generalization becomes particularly problematic when different aspects of the system need to be investigated separately or when multiple design iterations are required. The number of simulations necessary for tackling a design problem can quickly escalate into thousands, making the task laborious and time-consuming.

Recently, deep learning and neural networks, in particular, have been used extensively as surrogates to model the underlying physical phenomenon (\cite{jumper2021highly, carleo2019machine}). However, even state-of-the-art supervised machine-learning approaches encounter similar challenges as traditional experimental methods, requiring substantial input-output data at various fidelities to learn the underlying dynamics effectively. This large data requirement poses a significant hurdle, as obtaining such vast data can be arduous and resource-intensive (\cite{karniadakis2021physics}).

One potential approach to mitigate the need for an enormous amount of data is to incorporate the underlying physics into the learning procedure, thereby guiding the neural network based on physics principles as presented by \cite{raissi2019physics, karpatne2017physics, karpatne2017theory, karpatne2022knowledge}, among others. One popular class of methods that adopts this approach is physics-informed neural networks (PINNs) proposed by \cite{raissi2019physics}. PINNs are a form of semi-supervised learning where the boundary and initial conditions serve as input-output pairs while the solution is regularized by the governing partial differential equations (PDEs). However, several challenges for PINNs have emerged, including spectral bias (\cite{wang2022respecting}), shock learning (\cite{fuks2020limitations}), generalization with even slight changes in physical parameters and computational domain (\cite{kim2021dpm, kapoor2023neural, kapoor2023neuralparam}), and difficulties dealing with large coefficients (\cite{krishnapriyan2021characterizing, dekhovich2023ipinns, de2023operator}). 

One possible reason for such challenges encountered by vanilla PINNs is its causality-agnostic training. PINNs trained by gradient descent are implicitly biased towards first approximating PDE solutions at later times before resolving the initial conditions (\cite{wang2022respecting}), therefore profoundly violating temporal causality. Physical systems are known to possess an inherent causal structure. For instance, the deflection of the beam at any point in time is causally linked to the previous state of the system (deflection), the physical properties of the beam, and the external forces acting on it. This causality is a fundamental aspect of how the beam equations accurately model the behavior of beams in response to loads, making it a useful tool in engineering and physics. The PINN model could learn complicated solutions to PDEs when the causality is considered, enabling progressive sequential-time learning of the solution. 

Our work proposes to train PINN while respecting causality in the context of structural engineering, referred to as causal PINN hereafter. However, as we present in this work, even after employing causal PINN, the models lack generalizability to different initial conditions and computational domains, requiring each new problem to be solved from scratch. This limitation reverts the problem to the need for extensive simulations for each problem. To mitigate this issue, we propose employing transfer learning (TL) (\cite{niu2020decade}) in conjunction with causal PINN. The idea of transfer learning is to utilize the knowledge acquired from solving one problem in the form of trained model parameters to be utilized in a similar or related problem, accelerating the training process. 
 
We examine the application of PINNs on well-known Euler-Bernoulli and Timoshenko beam models on elastic foundations, specifically the Winkler foundation (\cite{lamprea2022beams, younesian2019elastic}). Numerical experiments reveal that the standard PINN approach struggles to capture the underlying physics accurately, particularly for large space-time domains, even in cases with closed-form analytical solutions. This paper proposes a novel approach to simulate beams on elastic foundations using the Euler-Bernoulli and Timoshenko theories, employing a transfer learning-based causal PINN framework to conduct comprehensive experiments. Specifically, transferring knowledge from one initial condition to another, handling noisy initial conditions, transferring knowledge for beams of different lengths, and systems with significant time dependencies are addressed. The primary contributions of this research paper are as follows:
 
A causality-respecting PINN loss function addresses the aforementioned limitations and effectively enforces the relevant physics. However, implementing this modified causal loss function requires a denser neural network with more parameters. Considering the importance of various factors in engineering structure design and the impracticality of simulating every instant, transfer learning is proposed within the causal PINN architecture. By incorporating transfer learning, the parameters of the previously trained model are leveraged to initialize and train new models. Consequently, this reduces the computational burden and enables faster convergence for subsequent tasks, improving the efficiency of simulating the dynamics of beams on elastic foundations.

The rest of the paper is structured as follows: Section~\ref{sec2} presents related works to this manuscript. Section~\ref{sec3} provides a detailed discussion of vanilla and causal PINN. Section~\ref{sec4} introduces the proposed framework of fusing transfer learning with causal PINN to train different models. Section~\ref{sec5} presents the numerical experiments results, showcasing the effectiveness of our methodology in addressing challenging beam problems where the vanilla PINN (\cite{raissi2019physics}) and self-adaptive PINN (\cite{mcclenny2023self}) formulation fails. Finally, the main findings are summarized, and conclusions drawn from this study are presented in Section~\ref{sec6}.

\section{Related works}\label{sec2}
This section outlines the pertinent studies conducted within the domain of transfer learning-driven PINNs, causal PINNs, and physics-informed methodologies for the simulation of beam models.

Applying transfer learning within PINNs has garnered significant attention (\cite{goswami2020transfer}). Notably, \cite{li2023physics} predicted laser deposition temperature fields accurately without labeled data, using physical losses and transfer learning. In another work, \cite{liu2022temperature} utilized transfer learning for accurate temperature field inversion with limited observations, employing a PINN and optimal position selection. \cite{roy2023data} developed a multi-objective loss function and transfer learning for accurate elastoplastic solid mechanics solutions through PINN. In a different study, \cite{haghighat2023constitutive} proposed a transfer learning-based PINN framework for efficient stress-strain constitutive modeling. While our research aligns with the fundamental principle of leveraging transfer learning, a distinguishing aspect lies in our consideration of causality during the training of the models. 

In the literature, research has been conducted to enforce causality in the PINN framework without incorporating transfer learning (\cite{mu2023pirt}). In another work, \cite{penwarden2023unified} proposed a causal framework incorporating transfer learning to simulate time-dependent PDEs. Although our work shares similar ideas of incorporating causality and utilizing transfer learning within the PINN framework, we employ transfer learning to train distinct models under diverse conditions. Conversely, \cite{penwarden2023unified} employs transfer learning within a particular problem by segmenting the domain into multiple subdomains and leveraging insights from one subdomain to another, employing the concepts of domain decomposition and PINNs (\cite{jagtap2021extended}).

Recently, beam simulations have concentrated on physics-informed methodologies, largely omitting the considerations of causality and transfer learning. Noteworthy works include \cite{bazmara2023physics}, which utilized PINNs for estimating nonlinear bending behavior within a confined domain. Similarly, \cite{10255379, kapoor2023ML} delved into applying PINNs for the system of beam models and moving load problems, albeit within the limited domain confines. \cite{lee4362563physics} introduced a spatio-temporal PINN tailored for analyzing the dynamics of cantilever beams. In \cite{xu2023transfer}, a self-adaptive PINN framework capable of accommodating varying load conditions is presented. Additionally, \cite{fallah2023physics} sought to enhance predictions by incorporating supplementary data, all still constrained within the confined domain bounded by the capabilities of PINNs. This work aims to enhance the potential of physics-informed methodologies for simulating beam dynamics.

\section{Vanilla and causal PINN} \label{sec3}
This section is structured into two subsections. First, we provide an overview of the architecture of the vanilla PINN (\cite{raissi2019physics}). Second, a modification in the PINN loss function leading to the incorporation of causality in the PINN loss function, as proposed by \cite{wang2022respecting}, is presented.

\subsection{Vanilla PINN} 
Recently, PINNs have been widely used for solving PDEs across diverse domains, including but not limited to works by \cite{chen2023tgm, zobeiry2021physics, shen2021physics}. PINNs are based on deep neural network (DNN) architecture, and the idea of PINN is to incorporate physical knowledge in the loss function of DNN. The loss function consists of two terms - a data term and a physics term. The data term ensures that the neural network fits the provided data points, while the physics term enforces the PDE constraints. Here, the data term refers to the value of the quantity of interest at initial and boundary points. Minimizing the data term amounts to measuring the discrepancy between the predicted solution of the PINN and the measured data points. The physics term incorporates the PDE constraints into the loss function, evaluating the differential operator of the PDE using automatic differentiation (\cite{paszke2017automatic}). The resulting equation is then included as a penalty term in the loss function. To elucidate these terms, we consider an abstract PDE as,
\begin{equation}
\label{eq1}
   D(u(x, t, k)) = f(x, t), \quad (x, t) \in \mathcal{D} \times \mathcal{T} \\
\end{equation}
where $D$ is the differential operator, $\mathcal{D}$ is the spatial domain, and $\mathcal{T}$ is the temporal domain. The unknown solution is $u$ depending on independent space ($x$) and time ($t$) variables. A constant parameter is $k$, and $f$ is the source term. To ensure the uniqueness of the solution, appropriate initial and boundary conditions are necessary for the considered PDE.
\begin{equation}
\label{eq2}
\begin{gathered}
u(x, 0) = g(x), \quad (x, 0) \in \mathcal{D} \times \Gamma \\
 u(x_\mathrm{b}, t) = \bar{g}(x_\mathrm{b}, t), \quad (x_\mathrm{b}, t) \in \Omega \times \mathcal{T}
\end{gathered}
\end{equation}
here, $g(x)$ and $\bar{g}(x_\mathrm{b}, t)$ are the initial and boundary conditions, respectively. The initial temporal region and spatial boundary are $\Gamma$ and $\Omega$, respectively. The loss function of PINNs is defined as follows 
\begin{equation}
\label{eq3}
    L(\mu) = \lambda_1L_\mathrm{PDE}(\mu) + \lambda_2L_\mathrm{IC}(\mu) + \lambda_3L_\mathrm{BC}(\mu)
\end{equation}
here, $\mu$ represents the trainable network parameters. The individual loss terms weighted by the hyperparameters $\lambda_i$, $i=1, 2, 3$, are defined as,
\begin{equation}
\label{eq4}
\begin{gathered}
L_\mathrm{PDE}(\mu) = \frac{1}{N_\mathrm{int}}\sum_{n=1}^{N_\mathrm{int}} ||D(u^{*}(x^{(n)}, t^{(n)}, k)) - f(x^{(n)}, t^{(n)})||^p 
\end{gathered}
\end{equation}
The loss terms for initial and boundary conditions in ~\eqref{eq3} are defined as follows,
\begin{equation}
\begin{gathered}
\label{eq5}
L_\mathrm{IC}(\mu) = \frac{1}{N_\mathrm{i}}\sum_{n=1}^{N_\mathrm{i}} ||u^{*}(x^{(n)}, 0) - g(x^{(n)})||^p \\
L_\mathrm{BC}(\mu) = \frac{1}{N_\mathrm{b}}\sum_{n=1}^{N_\mathrm{b}} ||u^{*}(x^{(n)}_\mathrm{b}, t^{(n)}) - \bar{g}(x_\mathrm{b}^{(n)}, t^{(n)})||^p  
\end{gathered}
\end{equation}
here, $N$ is the total number of training points, which is the sum of interior training points ($N_\mathrm{int}$), initial training points ($N_\mathrm{i}$), and boundary training points $N_\mathrm{b}$. The approximation of $u$ by the neural network is denoted by $u^{*}$. Training with $L^{2}$-norm amounts to $p$ $=$ $2$. The primary objective is minimizing~\eqref{eq3} and obtaining optimal parameters ($\mu$). These optimized parameters are then utilized for predicting the PDE solution $u(x, t), \forall (x, t) \in \mathcal{D} \times \mathcal{T}$. 

\subsection{Causal PINN}
This subsection presents causal PINN, modifying the PINN loss function (\cite{wang2022respecting}). The notion of causal PINNs is inspired by traditional numerical methods for solving differential equations that prioritize resolving the solution at lower times before approximating the solution at higher times. The modification in the loss function pertains to the PDE term $L_\mathrm{PDE}(\mu)$, while the initial $L_\mathrm{IC}(\mu)$ and boundary $L_\mathrm{BC}(\mu)$ loss terms remain unchanged. The causal PDE loss term $L_\mathrm{PDE}(\mu)$ is defined as
\begin{equation}
\begin{gathered}
\label{eq6}
L_\mathrm{PDE}(\mu) = \sum_{i=1}^{N_\mathrm{t}} w_i L_\mathrm{PDE}(t_i, \mu) \\
w_i = e^{-\epsilon \sum_{k=1}^{i-1} L_\mathrm{PDE}(t_k, \mu)}, \quad i=2,3, \ldots N_\mathrm{t}
\end{gathered}
\end{equation}

Here, $N_\mathrm{t}$ is the number of timesteps in which the computational domain has been divided. The causality hyperparameter $\epsilon$ controls the steepness of the weights. The modification introduces a weighting factor, $w_i$, for loss at each time level $t_i$. The weight $w_i$ depends on the accumulated PDE loss up to time $t_i$. The weights are adjusted to prioritize the fully resolved solution at lower time levels by exponentiating the negative of this accumulated loss. To summarize, the modified loss function for causal PINN could be written as
\begin{equation}
\label{eq7}
L_\mathrm{PDE}(\mu) = \frac{1}{N_t} \sum_{i=1}^{N_t} e^{-\epsilon \sum_{k=1}^{i-1} L_\mathrm{PDE}(t_k, \mu)} L_\mathrm{PDE}(t_i, \mu)
\end{equation}

In the following section, the proposed transfer learning method is presented along with the underlying motivation.

\section{Transfer learning for causal PINN} \label{sec4}

\begin{figure*}[t]
\begin{center}
\includegraphics[width=0.90\textwidth]{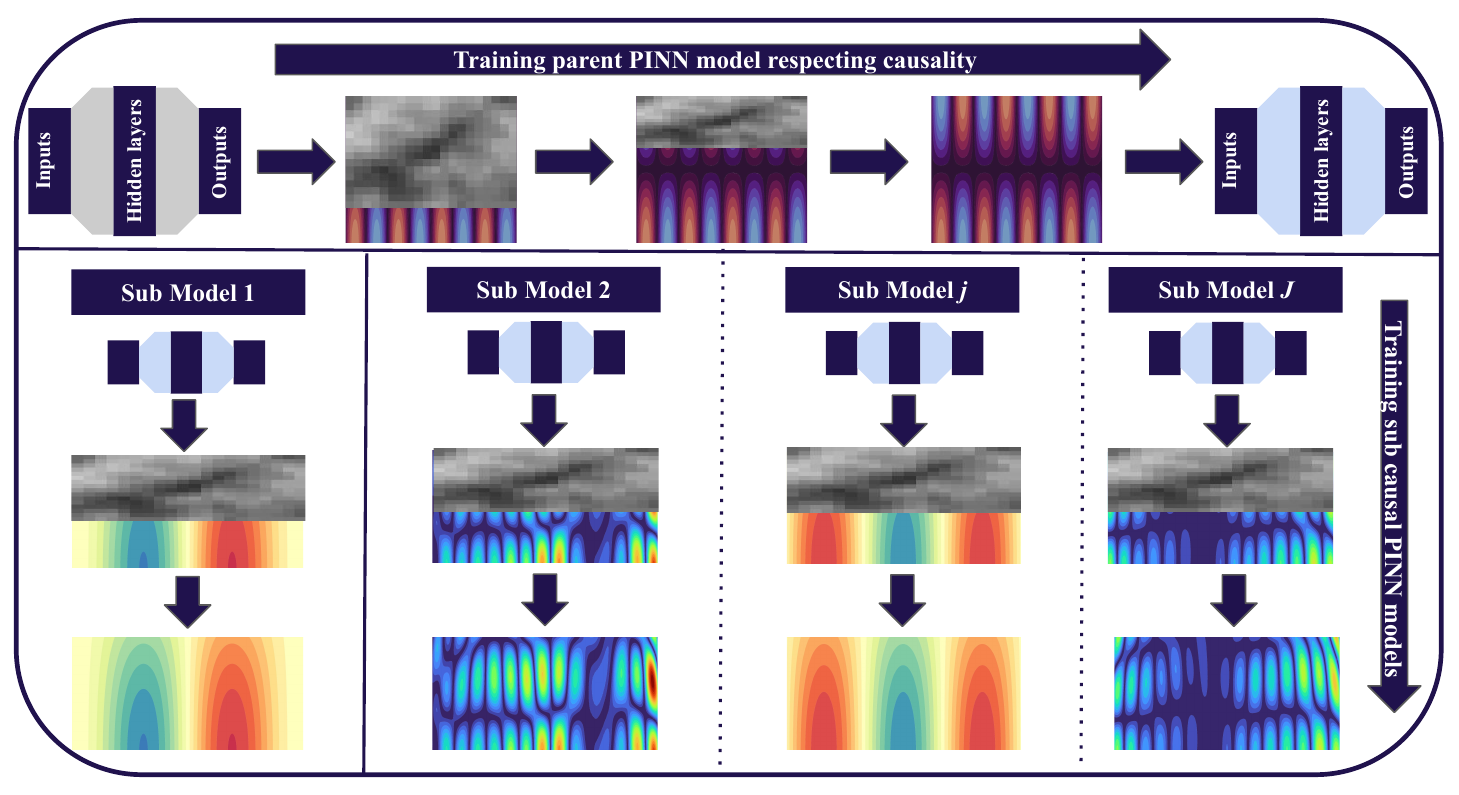}
\caption{\textbf{Proposed Transfer Learning Framework in Causal PINN:} The top horizontal block outlines the training process of the causal PINN for the parent model, which is the primary beam model under consideration. The model parameters, initialized using Xavier initialization within the first neural network, undergo training while adhering to causality. The resulting parameters of the trained model serve as the initialization for $J$ subsequent tasks shown by the bottom vertical blocks ($1 \leq j \leq J$). These tasks pertain to different initial conditions and extensions of both spatial and temporal domains. The training of these subtasks is also performed to adhere to causality.}
\label{fig:proposed_architecture}
\end{center}
\end{figure*}

Several factors are crucial for designing an engineering structure, and solving the problem for each case is important. However, training the neural network for every case is time-consuming and laborious. Here, we propose to utilize transfer learning for beam problems on the Winkler foundation. The idea is to train the parent beam model for one case, for instance, to train an Euler-Bernoulli beam for a specific initial condition and then utilize the parameters for different initial conditions. The aim is to reduce the training time for the transfer learning case compared to the case without transfer learning. This reduction in computational time in terms of epochs is done by utilizing the previously trained model parameters and using them as initialization for subsequent cases. 

The proposed approach incorporates transfer learning for different scenarios for the same physical beam equation. Fig.~\ref{fig:proposed_architecture} visually demonstrates the steps: initially, the parent model is trained using causal PINN for a significant number of epochs ($n_1$). Subsequently, the trained parameters are utilized as an initialization for the training of other problems of the physical equation with different initial conditions or for an extended domain, which is trained for a reduced number of epochs ($n_2$), where $n_2$ $<<$ $n_1$, reducing the computational cost of training the model again from the start. The step-by-step illustration is provided in Fig.~\ref{fig:proposed_architecture}.

In Fig.~\ref{fig:proposed_architecture}, the top horizontal block illustrates the training of causal PINN for the parent model, specifically the primary beam model, either the Euler-Bernoulli or the Timoshenko beam model. The model parameters, generated using Xavier initialization (\cite{glorot2010understanding}) for the initial neural network, undergo a training process adhering to the causal loss function. This training involves the resolution of solutions at lower times prior to approximating at higher times, as shown by the snapshots of the resolved solution. As the number of epochs increases, the model prediction at higher time levels improves only when the solution at lower time levels has been resolved up to a certain accuracy. The resulting parameters from this training serve as the initialization for subsequent $j$ tasks presented by the bottom vertical blocks in Fig.~\ref{fig:proposed_architecture}. These tasks involve diverse initial conditions and extensions of both spatial and temporal domains. Notably, the training of these subtasks is also performed by minimizing the loss terms ~\eqref{eq7} and ~\eqref{eq5} in the loss function ~\eqref{eq3}, ensuring a coherent and principled transfer learning framework.

\begin{figure*} 
\centering
\subcaptionbox{}{\includegraphics[width=0.24\columnwidth]{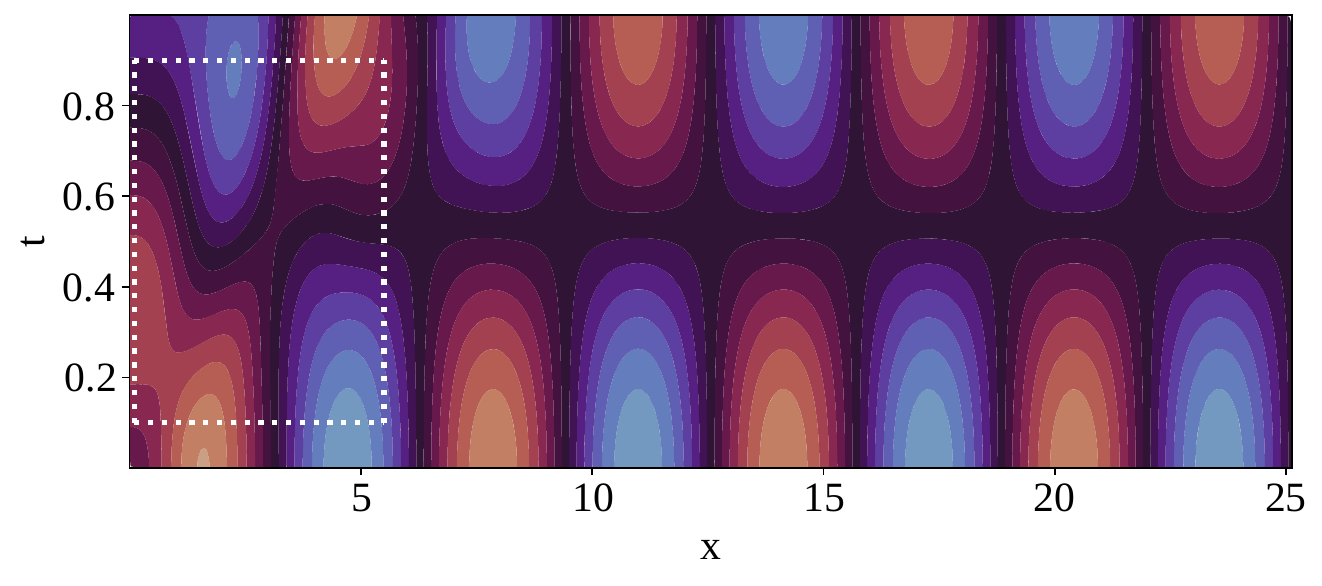}}
\subcaptionbox{}{\includegraphics[width=0.24\columnwidth]{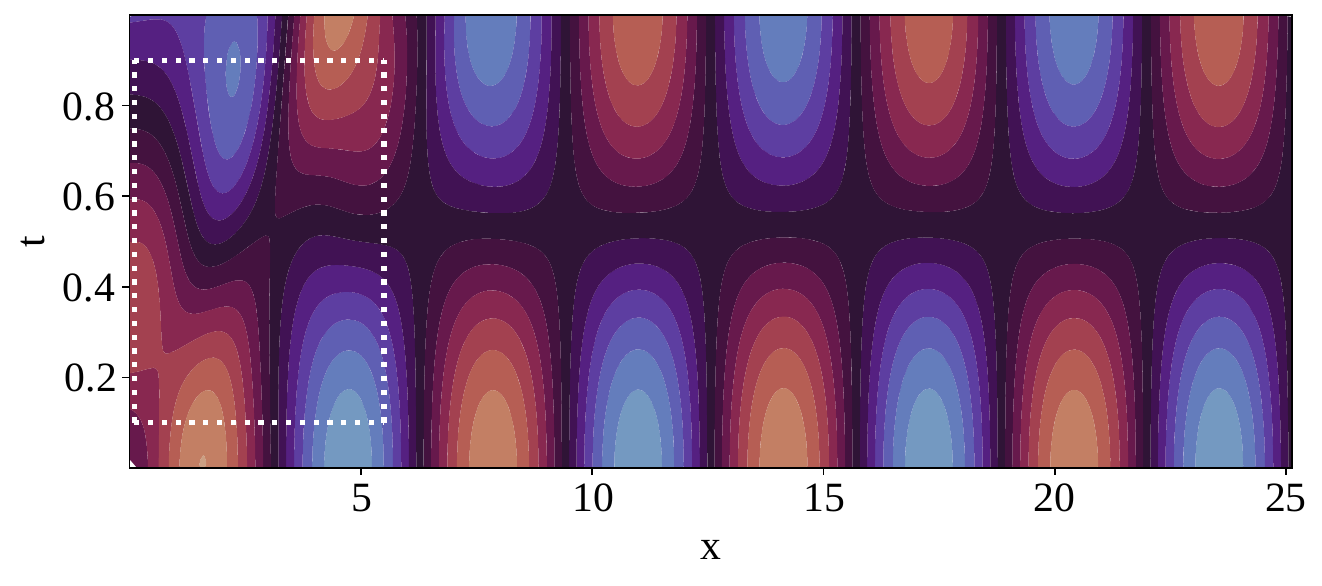}}
\subcaptionbox{}{\includegraphics[width=0.24\columnwidth]{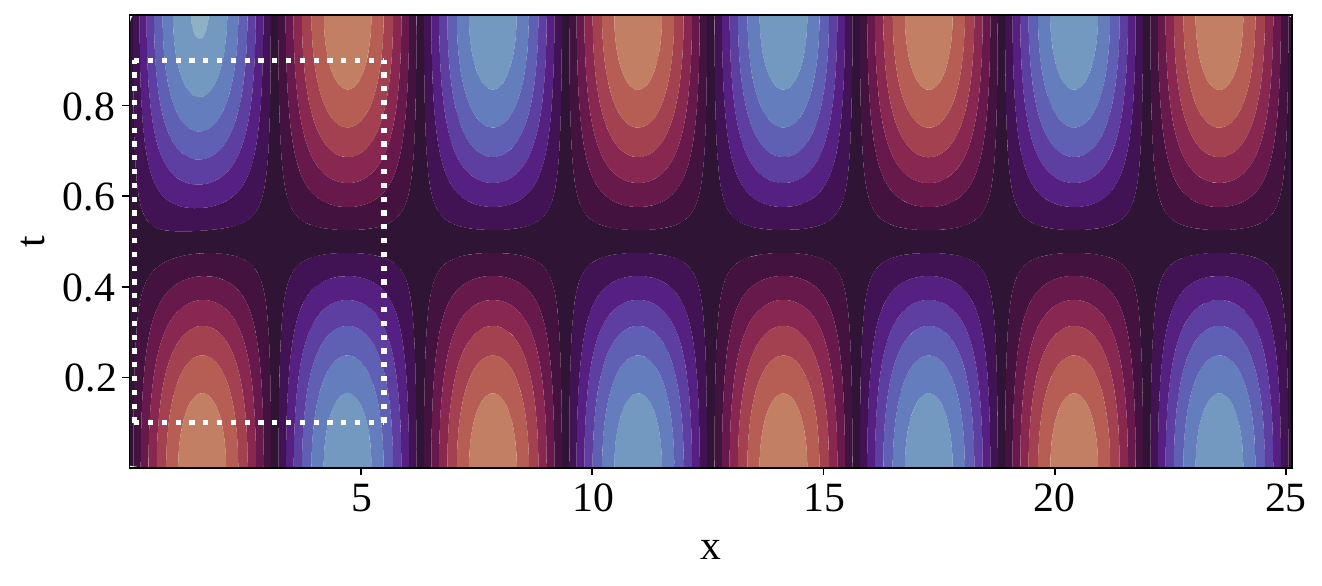}}
\subcaptionbox{}{\includegraphics[width=0.24\columnwidth]{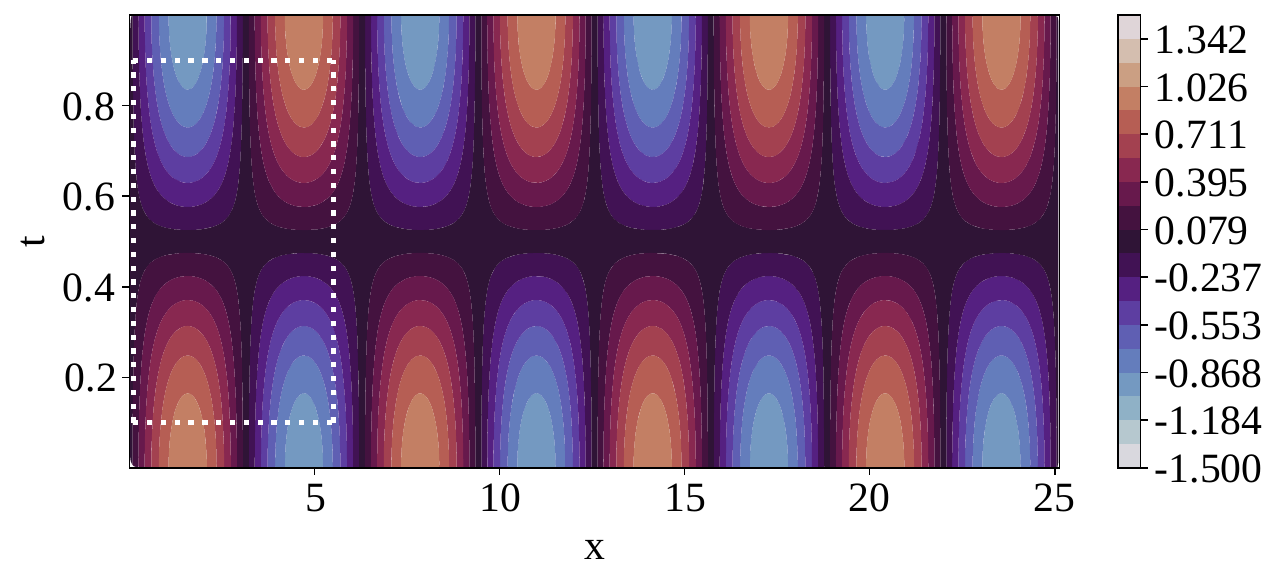}}
\caption{Euler-Bernoulli beam displacement on the Winkler foundation \textbf{(a.)} Predicted solution using PINN \textbf{(b.)} Predicted solution using SA-PINN  \textbf{(c.)} Predicted solution using causal PINN \textbf{(d.)} Reference solution}
\label{fig3}
\end{figure*}

In the next section, we perform a series of experiments to demonstrate the efficacy of the proposed framework.

\section{Numerical experiments} \label{sec5}
This section presents the numerical experiments for simulating the dynamics of the Euler-Bernoulli and Timoshenko beam models using the proposed framework. The proposed method is compared with vanilla PINN and self-adaptive PINN (SA-PINN) proposed by \cite{mcclenny2023self} and utilized in several works, for instance, \cite{ding2023self} to improve the performance of PINNs. In addition, leveraging transfer learning, several other experiments are performed for noisy data, different initial conditions, and extrapolation in both spatial and temporal domains for the beam models.

The experimental setup involves first simulating the parent case and utilizing the trained parameters for various subcases. Specifically, transfer learning is employed for these subcases. The main model utilizes a neural network architecture comprising four hidden layers with 200 neurons each. The activation function employed is the hyperbolic tangent (tanh), and the limited-memory Broyden-Fletcher-Goldfarb-Shanno (LBFGS) optimizer is utilized with a learning rate of $0.1$. The parent model is trained for a total of $10,000$ epochs. Within the causal-respecting PINN function, the causality hyperparameter ($\epsilon$) is set to $5$ and the number of timesteps $N_t$ is taken to be 100. During the training process, $N_\mathrm{i} = 500$ initial points, $N_\mathrm{b} = 1000$ boundary points, and $N_\mathrm{int} = 10,000$ interior points are considered. The weight hyperparameters $\lambda_{1}$, $\lambda_{2}$ and $\lambda_{3}$ are taken to be $1$ each. The selected evaluation metric is the $L^2$ relative error percentage ($\mathcal{R}$) defined as
\begin{equation}
\begin{aligned}
\label{eq8}
\mathcal{R} = \frac{||u^{*} - u ||_{2}}{||u||_{2}} \times 100
\end{aligned}
\end{equation}
where $u^{*}$ is the approximated PDE solution by the neural network, and $u$ refers to the ground truth. We utilize the trained parameters $(\mu)$ of the main model as initialization for training the subcase neural networks for only $1500$ epochs, achieving the same level of accuracy as the main model.

\begin{figure*} 
\centering
\includegraphics[width=0.48\columnwidth]{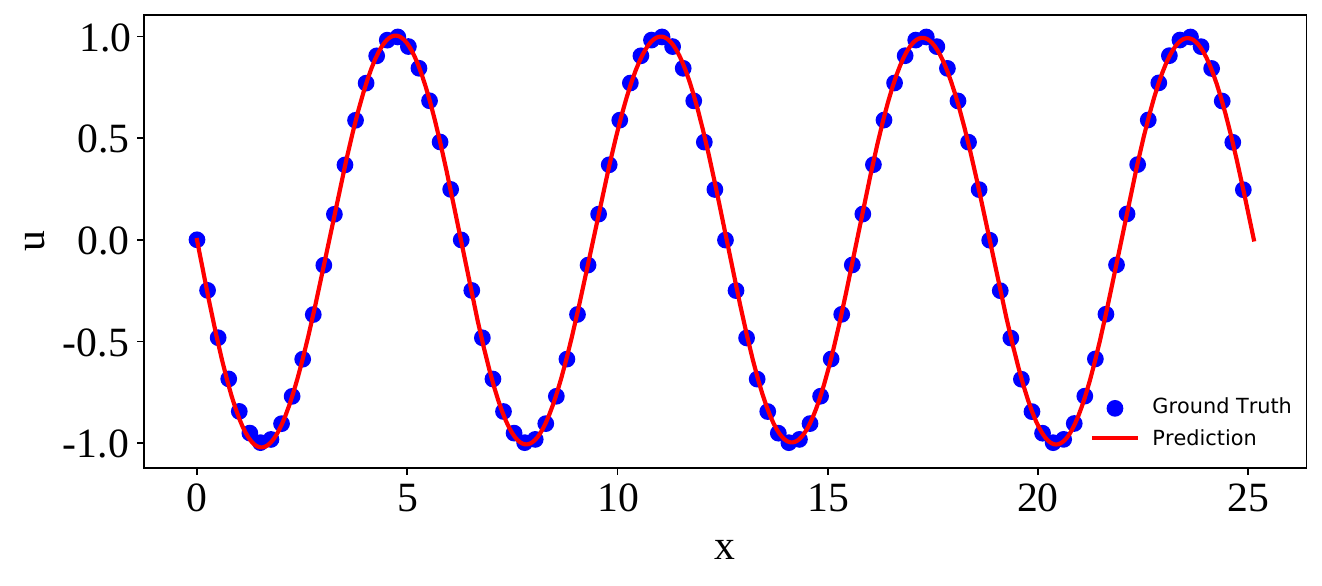}
\includegraphics[width=0.48\columnwidth]{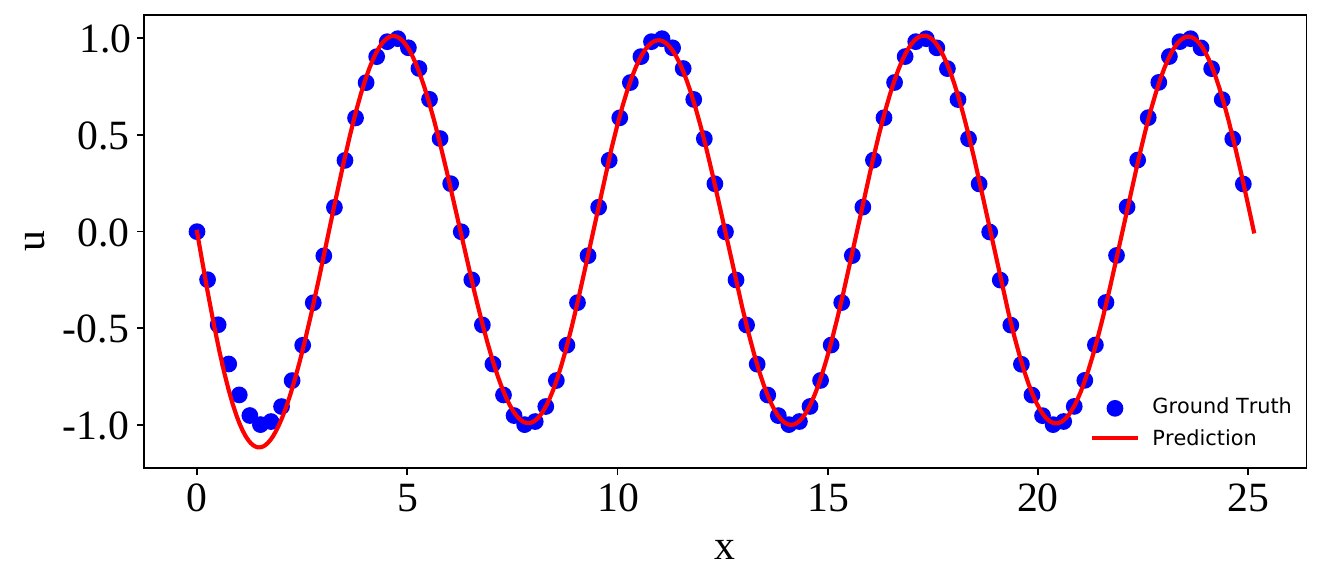}
\caption{Euler-Bernoulli beam equation on the Winkler foundation with noise in the initial condition of displacement of the beam  \textbf{Left:} Predicted solution at final time ($t = 1$) with $10\%$ Gaussian noise; \textbf{Right:} Predicted solution at final time ($t = 1$) with $20\%$ Gaussian noise}
\label{fig4}
\end{figure*}

\subsection{Euler-Bernoulli beam}
The Euler-Bernoulli beam model is a mathematical framework used to analyze the behavior of beams when subjected to loads. It is derived from the three-dimensional elasticity theory or through principles such as Newton's second law or the generalized Hamiltonian Principle (\cite{ochsner2021classical}). The model assumes certain simplifications: it neglects the effects of rotary inertia and transverse shear deformations. The Euler-Bernoulli beam equation describes the behavior of a beam subjected to bending (Fig.~\ref{fig1}). When the beam is supported on a Winkler foundation, representing an elastic foundation, the Euler-Bernoulli beam equation is modified to account for the interaction between the beam and the foundation. This modified equation considers the stiffness of the foundation and its influence on the behavior of the beam. The mathematical model of a simply supported Euler-Bernoulli beam on a Winkler foundation is described by (\cite{younesian2019elastic})
\begin{equation}
\label{eq9}
u_\mathrm{tt} + u_\mathrm{xxxx} + p(x, t) = f(x, t), \quad x \in [0, 8\pi], t \in [0, 1]
\end{equation}
where $u$ represents the vertical displacement of the beam. $u_\mathrm{tt}$, and $u_\mathrm{xxxx}$ represent the two times partial derivative of $u$ with respect to $t$, and four times partial derivative with respect to $x$, respectively. The loading on the beam is defined by $f(x, t) = (2 - \pi^2)\sin{(x)}\cos(\pi t)$. The initial and boundary conditions are given as
\begin{equation}
\begin{gathered}
\label{eq10}
u(x, 0) = \sin(x), \quad u_\mathrm{t}(x, 0) = 0 \\
u(0, t) = u(8\pi, t) = u_\mathrm{xx}(0, t) = u_\mathrm{xx}(8\pi, t) = 0
\end{gathered}
\end{equation}

The foundation reaction, $p(x,t)$, assumes that the reaction at every location is proportional to the displacement at a particular location, and the springs are linear and independent, as described in~\eqref{eq9}. The reaction force of the foundation is given by $p(x, t) = k u(x, t)$, where $u(x, t)$ is vertical displacement and $k$ is the stiffness of linear springs. The exact solution for this problem is given by $u(x, t) = \sin(x)\cos(\pi t)$.

Solving~\eqref{eq9}, one can determine the vertical displacement of the beam at any point along its length and other important quantities of interest, such as bending moments and beam acceleration. These quantities help engineers understand how the beam will perform structurally and ensure it meets the desired design criteria. By calculating the displacement, engineers can check whether the beam deflects within acceptable limits under the applied loads.

\begin{table}[htbp]
\caption{Euler-Bernoulli beam: $\mathcal{R}$ at $t=1$ for $k = 1$}
\centering
\begin{tabular}{cccc}
\toprule
 & PINN & SA-PINN & Causal PINN \\
\midrule
$\mathcal{R}$ & 5.33732 & 5.15410 & 0.03618 \\
\bottomrule
\end{tabular}
\label{tab1}
\end{table}
We simulate~\eqref{eq9} with three different methods to establish that incorporating causality provides more accuracy in the predicted solution than vanilla PINN and the SA-PINN for our problem. The results presented in Table~\ref{tab1} indicate that vanilla PINN and SA-PINN provide less accurate displacement predictions at $t = 1$ for the Euler-Bernoulli equation for stiffness $k=1$. In contrast, causal PINN yields more accurate displacement predictions as the relative percent error is $0.036$. This observation is further supported by the findings depicted in Fig.~\ref{fig3} (a), (b), which demonstrates that PINN and SA-PINN models are not accurate, particularly during the initial time, highlighted by the white rectangular box in Fig.~\ref{fig3}. However, this challenge is effectively overcome by incorporating a causality-respecting loss function (Fig.~\ref{fig3}(c)), which facilitates training the solution at lower time levels before training at higher times. 

The parameters from PINN and SA-PINN are not used subsequently to avoid incomplete or bad knowledge transfer. Only the trained parameters from the causal PINN formulation are transferred to the subsequent experiments presented in the next two subsections, fostering convergence by effectively reducing the training epochs.

\subsubsection{Noisy initial conditions}
This subsection presents the performance of the proposed method with noisy initial conditions. Initial conditions may not be perfectly known in real-world scenarios or contain uncertainties or noise. By learning displacements for noisy initial conditions, we can develop models that accurately represent the system's behaviour under such realistic conditions, allowing us to account for uncertainties and better understand the actual response of the system. To observe the dynamics of beam models under these conditions, we introduce Gaussian noise in the initial condition ranging from $5{\%}$ to $20{\%}$. The hyperparameter selection is the same as the main model, except for the number of epochs. With transfer learning, we perform $1500$ epochs instead of $10000$. 

Table~\ref{tab2} presents the results from $5{\%}$ to $20{\%}$ Gaussian noise levels in the initial conditions for the displacement of the beam with and without (w/o) using transfer learning. The proposed method predicts $u$ with less relative error percent. This prediction is significantly more accurate compared to the case without transfer learning. Also, Fig.~\ref{fig4} shows the results for $10{\%}$ and $20{\%}$ noise levels in the initial conditions for the displacement of the beam using transfer learning, demonstrating the computational efficiency of the proposed method. 

\begin{figure}[htbp]
\centerline{\includegraphics[width=0.5\columnwidth]{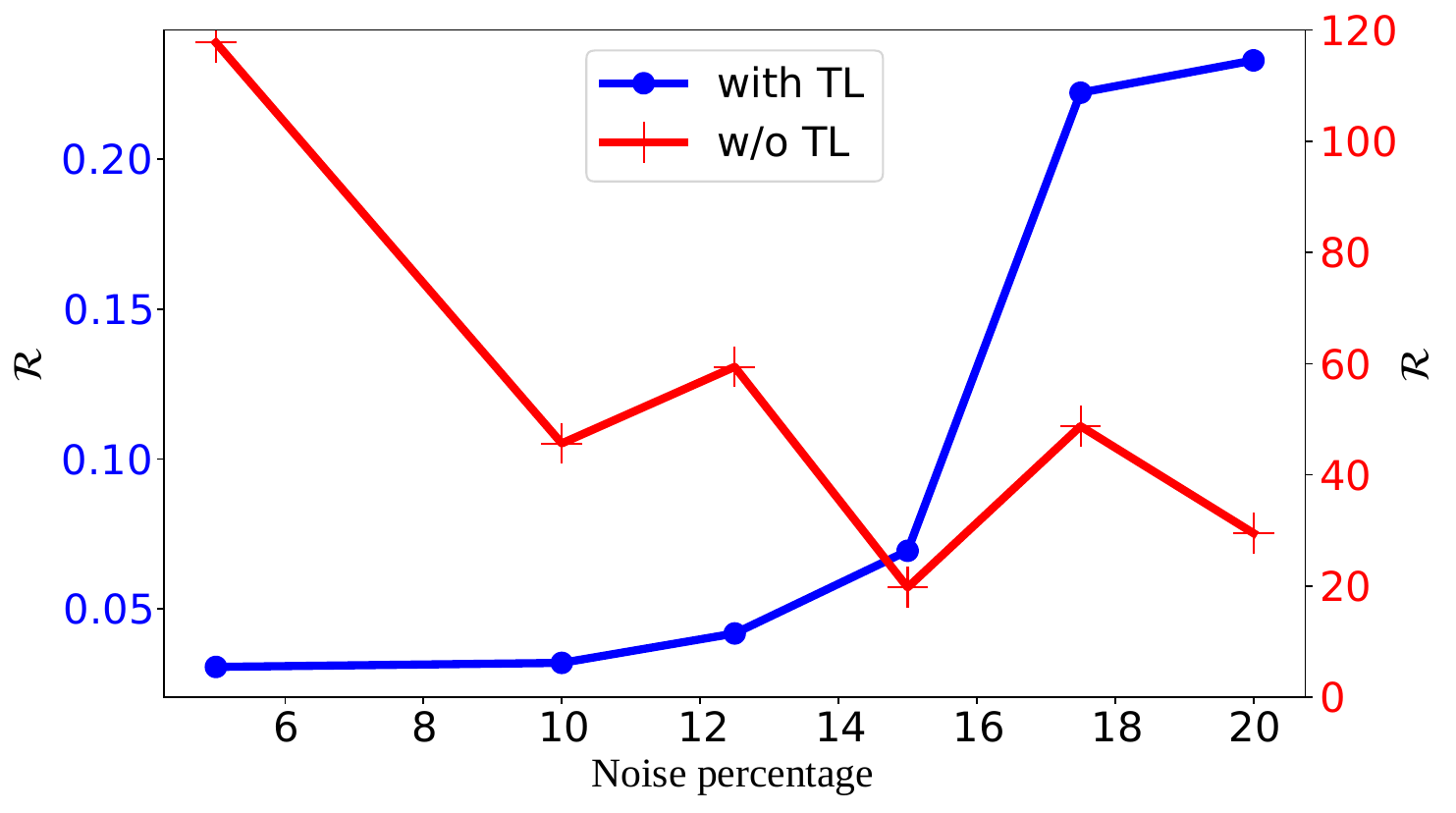}}
\caption{$\mathcal{R}$ vs noise percentage in the initial condition for the Euler Bernoulli beam for both approaches - with and without transfer learning.}
\label{fig5}
\end{figure}

Fig.~\ref{fig5} illustrates the comparison of relative error percentages concerning the noise percentage for both methods, one with transfer learning and the other without it. In the transfer learning scenario, it becomes apparent that an increase in the noise percentage results in a corresponding increase in the relative error percentage. When the subcases use the trained parameters for initialization, noise and error percentages exhibit a direct proportional relationship. However, in cases where trained parameters are not utilized, no discernible pattern emerges due to the non-convergence in minimizing the loss function. 

\begin{table}
\setlength{\tabcolsep}{2pt}
\centering
\scriptsize 
\caption{Euler Bernoulli Beam: $\mathcal{R}$ at $t=1$ for different percentages of noise in the initial condition. "TL" refers to transfer learning, and "w/o" refers to without. Abbreviations are used consistently for the following tables.}
\begin{tabular}{cccccccc}
\toprule
 & 5\% & 10\% & 12.5\% & 15\% & 17.5\% & 20\%  \\
\midrule
with TL & 0.03063 & 0.03198 & 0.04180 & 0.06937 & 0.222182 & 0.23296 \\
w/o TL & 117.7389 & 45.65849 & 59.42882 & 19.7473 & 48.75515 & 29.50691 \\
\bottomrule
\end{tabular}
\label{tab2}
\end{table}

\subsubsection{Different initial displacements and velocities}
In this section, we present the results of the Euler-Bernoulli beam for different initial conditions characterized by the change in initial displacements and velocities of the beam. Learning deflections for different initial conditions and force functions allows for generalization. Beams or structures can have varying initial conditions, such as different magnitudes, positions, or load distributions. By learning the deflections for a diverse set of initial conditions, we can develop models that capture the underlying patterns and behavior of the system, enabling accurate predictions for unseen or novel initial conditions. 

\begin{figure}[htbp]
\centerline{\includegraphics[width=0.5\columnwidth]{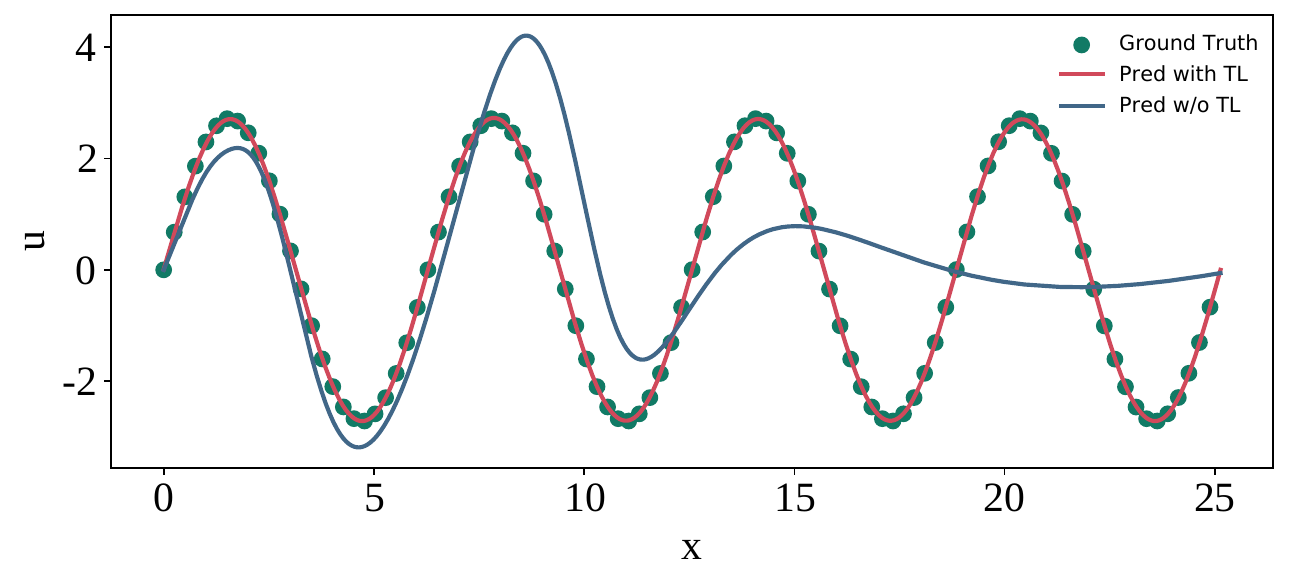}}
\caption{Euler-Bernoulli beam on the Winkler foundation for initial velocity for case 1: Causal PINN prediction at final time $t=1$ with and without transfer learning}
\label{fig6}
\end{figure}

Here,  we consider different initial conditions compared to the parent model. The initial conditions for this case are $u(x, 0) = a \sin(x)$ and $u_t(x, t=0) = a\sin(x)$. The analytical solution for the corresponding problem is $u(x, t) = a \sin(x)e^t$. We utilize the trained parameters of the Euler-Bernoulli beam model as an initialization for training this problem with different initial conditions considering $a=1, 2$ (representing case $1$ and case $2$ in Table~\ref{tab3}). The hyperparameters remain unchanged; the only change is the number of epochs, which is only 3000. Relative error percentages of displacement are presented in Table~\ref{tab3}, which shows a large difference in relative percent errors. From Fig.~\ref{fig6}, it is evident for the first case that the transfer learning approach achieves accurate predictions in a fewer number of epochs. 

\begin{table}[htbp]
\caption{Euler Bernoulli Beam: $\mathcal{R}$ at $t=1$ for different velocities}
\centering
\begin{tabular}{ccc}
\toprule
\textbf{$u^*$} & $\mathcal{R}$ (case 1) & $\mathcal{R}$ (case 2)  \\
\midrule
with TL & 0.00105 & 0.02188 \\
w/o TL & 70.72229  & 193.85024 \\
\bottomrule
\end{tabular}
\label{tab3}
\end{table}

\subsection{Timoshenko beam} 
The Timoshenko beam theory considers the shear deformation and rotational effects neglected in the Euler-Bernoulli beam equation (\cite{ochsner2021classical}). Hence, in addition to the quantity vertical displacement ($u$), Timoshenko's theory considers the cross-sectional rotation ($\theta$) as another unknown variable. The mathematical model for a beam resting on a Winkler foundation and subjected to an external load based on the Timoshenko beam theory is given as follows (\cite{younesian2019elastic})
\begin{equation}
\begin{aligned}
\label{eq11}
 \theta_\mathrm{tt} - \theta_\mathrm{xx} + (\theta - u_\mathrm{x}) = 0;  \\
u_\mathrm{tt} + (\theta - u_\mathrm{x})_\mathrm{x} + ku = h(x, t)    
\end{aligned}
 \end{equation}
where the symbols have their usual meaning, as in the case of the Euler-Bernoulli beam model. We consider $h(x, t)$ = $\cos(t)$ and the computational domain to be 
$x \in [0, 3\pi]$ and $t \in [0, 1]$. The supporting initial and boundary conditions are given as 

\begin{figure*} 
\centering
\subcaptionbox{}{\includegraphics[width=0.24\columnwidth]{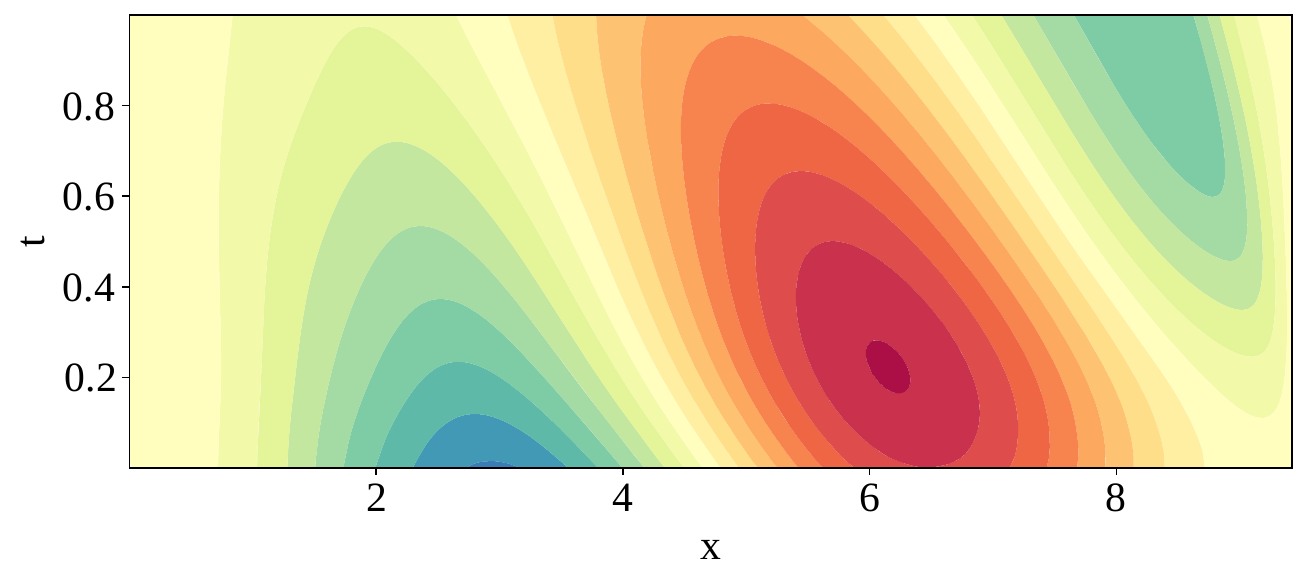}}
\subcaptionbox{}{\includegraphics[width=0.24\columnwidth]{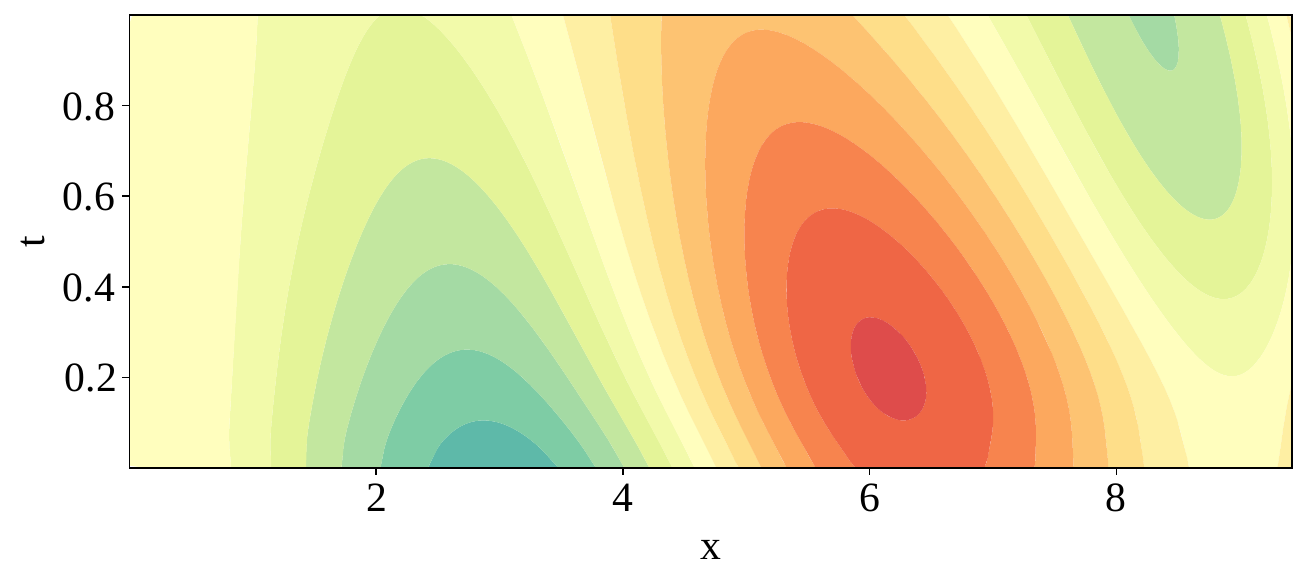}}
\subcaptionbox{}{\includegraphics[width=0.24\columnwidth]{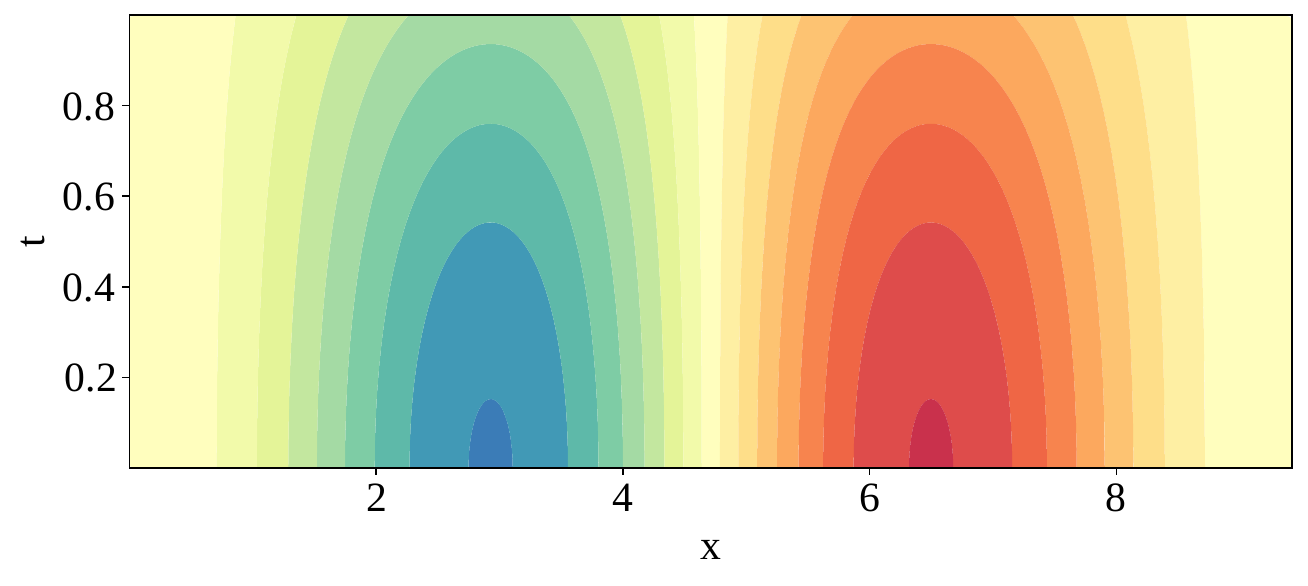}}
\subcaptionbox{}{\includegraphics[width=0.24\columnwidth]{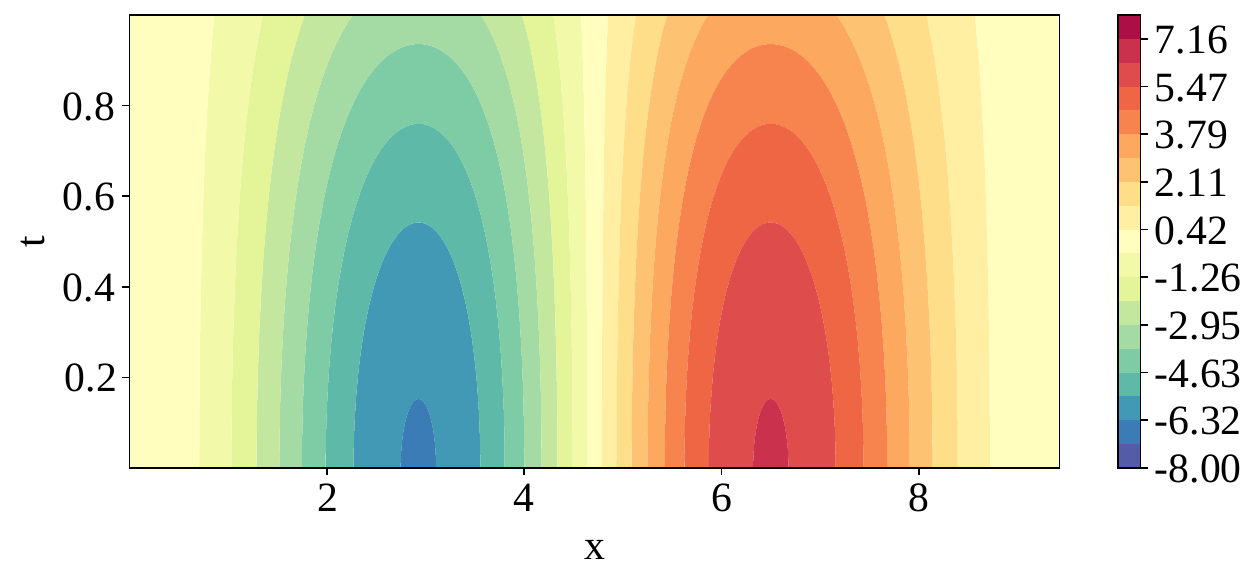}}
\subcaptionbox{}{\includegraphics[width=0.24\columnwidth]{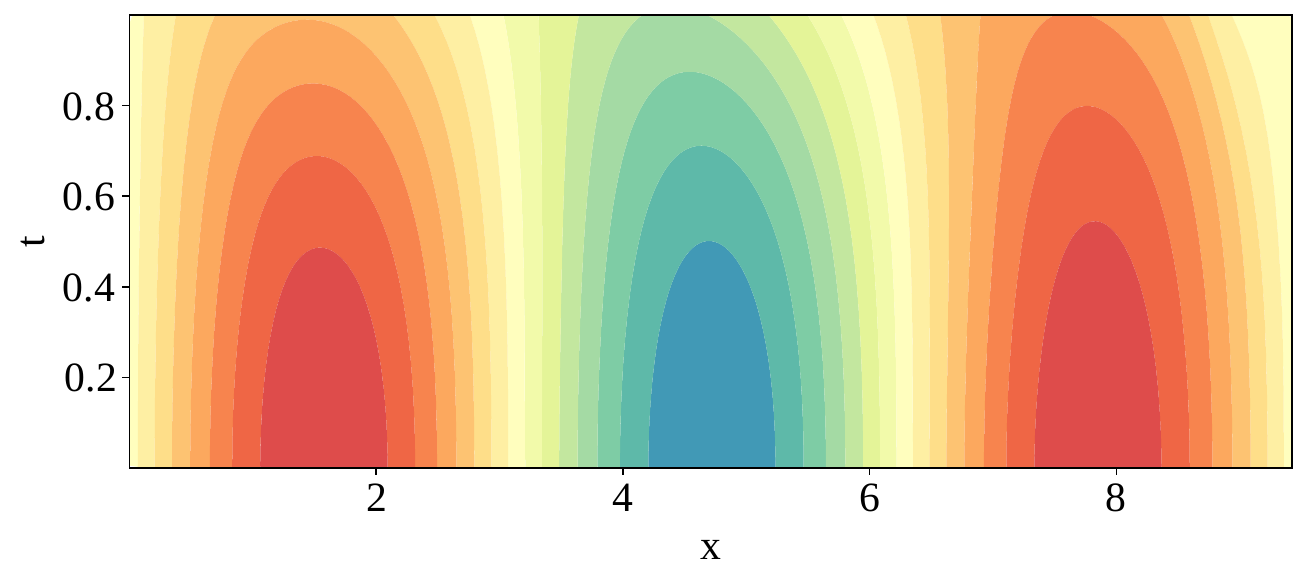}}
\subcaptionbox{}{\includegraphics[width=0.24\columnwidth]{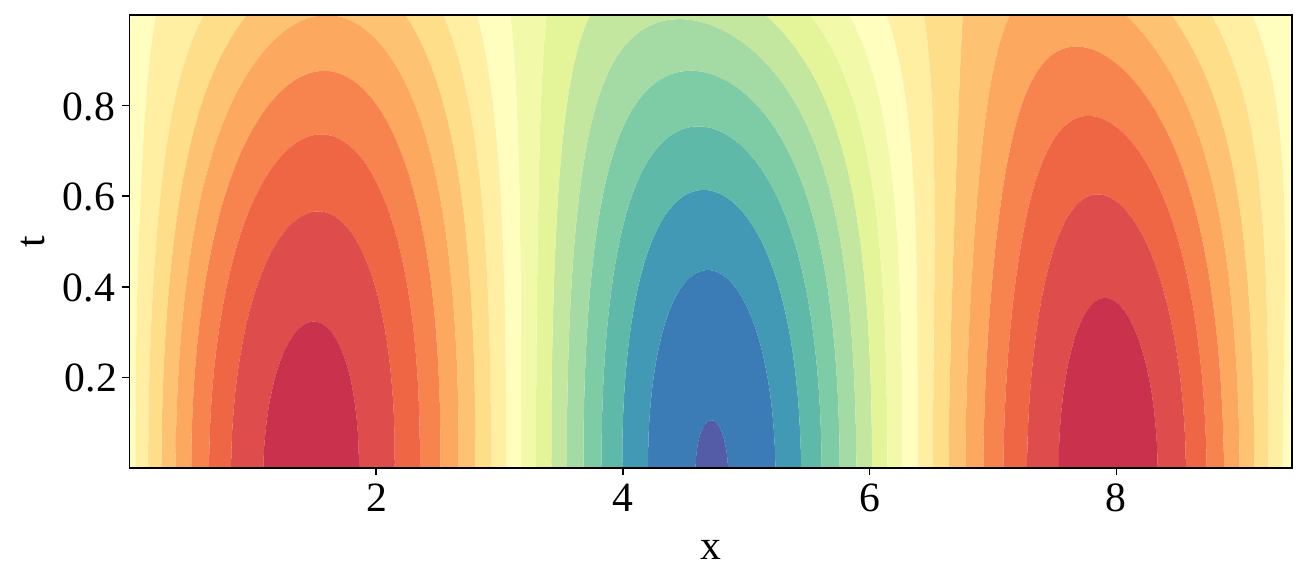}}
\subcaptionbox{}{\includegraphics[width=0.24\columnwidth]{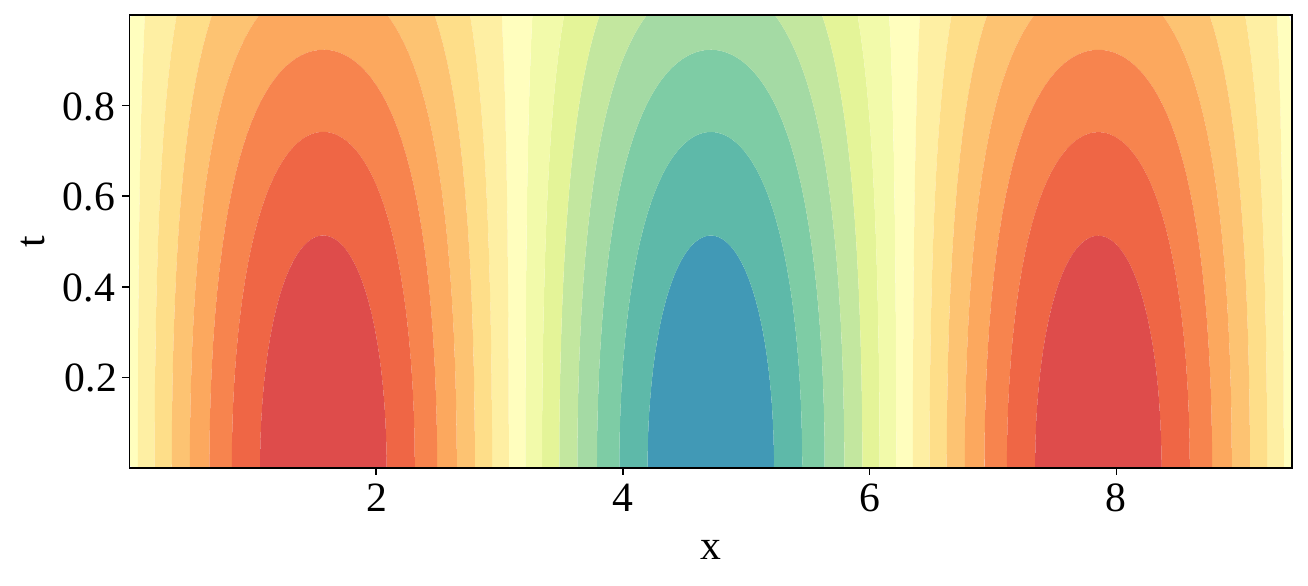}}
\subcaptionbox{}{\includegraphics[width=0.24\columnwidth]{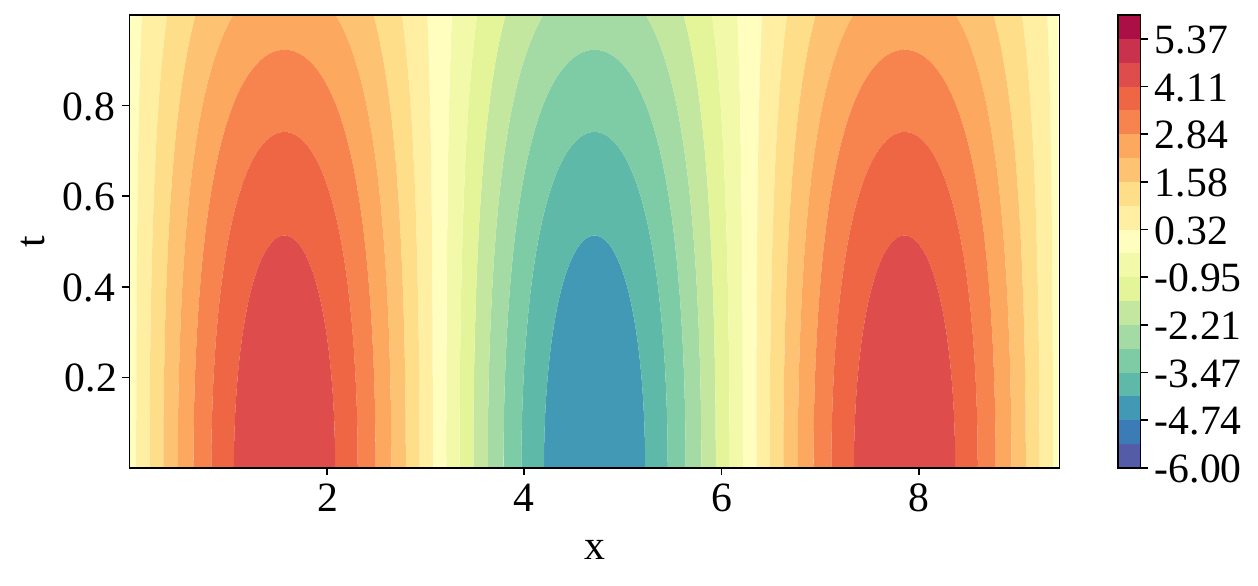}}
\caption{Timoshenko beam on the Winkler foundation \textbf{Top:} Predicted Displacement ($u^{*}$) \textbf{(a.)} Using PINN \textbf{(b.)} Using SA-PINN  \textbf{(c.)} Using causal PINN \textbf{(d.)} Reference solution \textbf{Bottom:} Predicted Rotation ($\theta^{*}$)  \textbf{(e.)} Using PINN \textbf{(f.)} Using SA-PINN  \textbf{(g.)} Using causal PINN \textbf{(h.)} Reference solution}
\label{fig7}
\end{figure*}

\begin{figure*} 
\centering
\includegraphics[width=0.48\columnwidth]{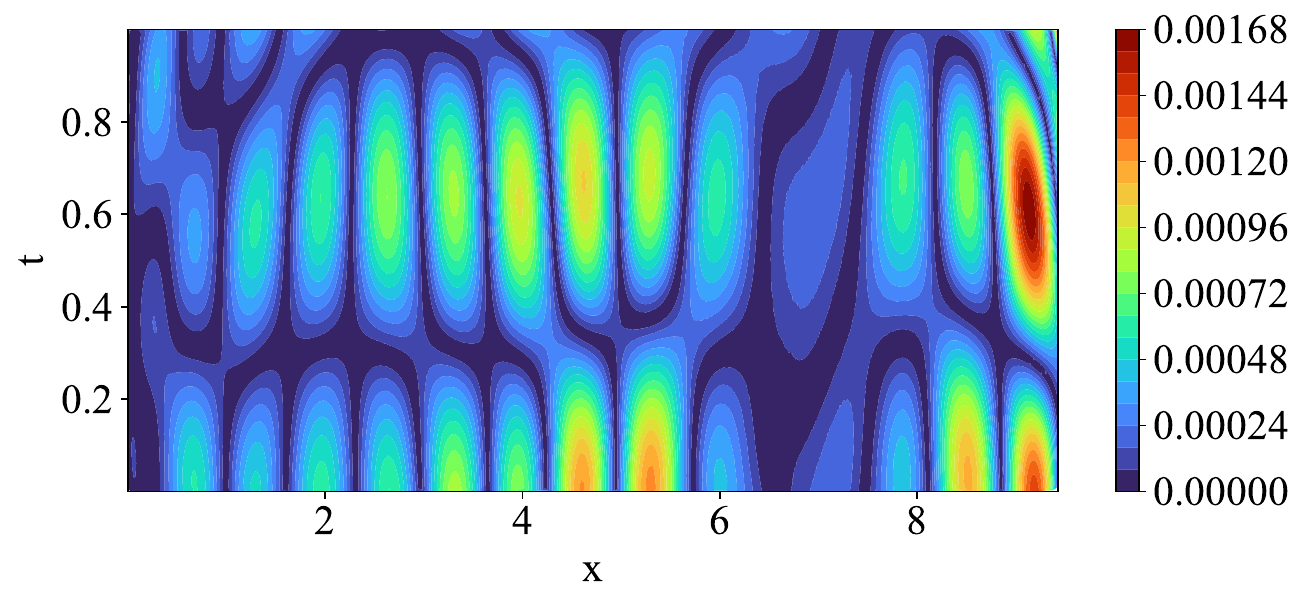}
\includegraphics[width=0.48\columnwidth]{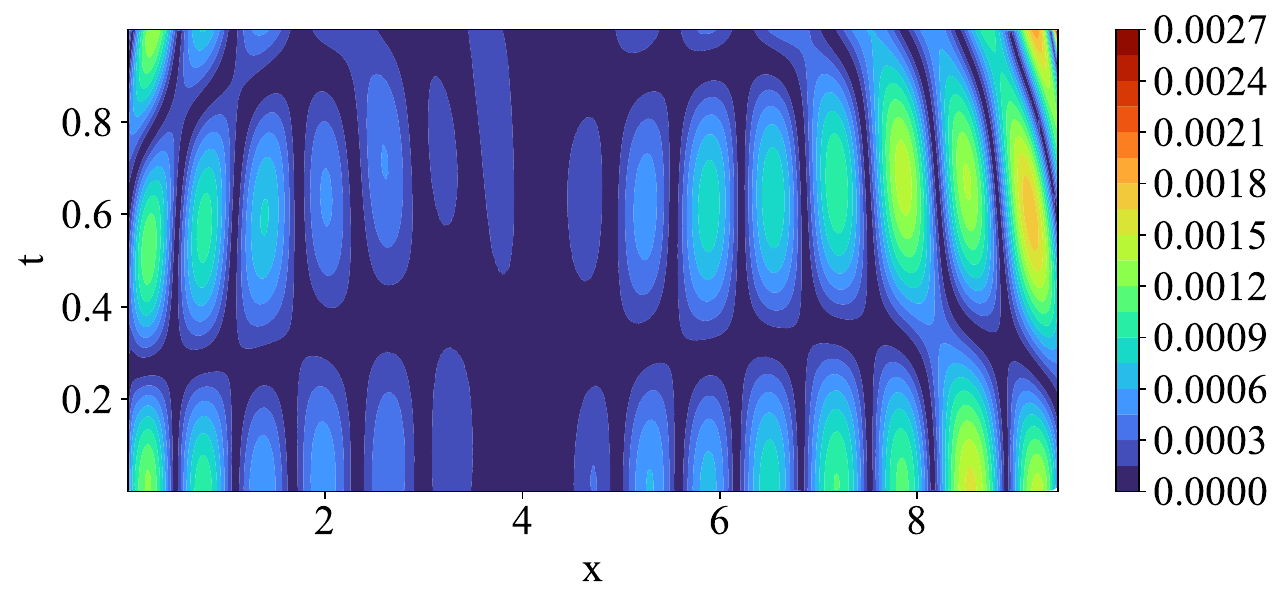}
\caption{Timoshenko beam on the Winkler foundation at final time $t = 1$. \textbf{Left:} Absolute error in predicting displacement ($|u - u^{*}|$); \textbf{Right:} Absolute error in predicting rotation ($|\theta - \theta^{*}|$)}
\label{fig8}
\end{figure*}

\begin{equation}
\begin{gathered}
\label{eq12}
    \theta(x, 0) = \frac{3\pi}{2}\cos(x) + \left(x - \frac{3\pi}{2}\right), \quad \theta_t(x, 0) = 0 \\
    u(x, 0) = \frac{3\pi}{2}\sin(x),\quad u_t(x, 0) = 0 \\
    \theta(0, t) = \theta(3\pi, t) = u(0, t) = u(3\pi, t) = 0
\end{gathered}
\end{equation}
 
The analytic solution for the rotation and vertical displacement is given as follows
\begin{equation}
\begin{gathered}
    \theta(x, t)= \left(\frac{3\pi}{2}\cos(x) + \left(x - \frac{3\pi}{2}\right)\right)\cos(t) \\
    u(x, t) = \frac{3\pi}{2}\sin(x)\cos(t)
\end{gathered}
\end{equation}

Solving the Timoshenko beam model~\eqref{eq11} -~\eqref{eq12} would help engineers obtain more accurate predictions of beam deflections and rotations, especially for beams with high aspect ratios or subjected to high shear forces. This accuracy is crucial for assessing structural integrity, ensuring compliance with design criteria, and preventing potential failures. 

Fig.~\ref{fig7} illustrates the predicted displacement and rotation throughout the entire space-time domain. Fig.~\ref{fig7}(c) and Fig.~\ref{fig7}(g) depict the displacement and rotation prediction using the causal PINN loss function. Fig.~\ref{fig7}(a-b) and Fig.~\ref{fig7}(e-f) depict the displacement and rotation prediction using vanilla PINN and SA-PINNs, respectively, illustrating its failure in prediction. In addition, Fig.~\ref{fig8} presents the absolute error in displacement and rotation resulting from the causal PINN loss function. The maximum error magnitude falls below $10^{-2}$, clearly indicating the accuracy of causal PINN.

\begin{table}[htbp]
\caption{Timoshenko Beam: $\mathcal{R}$ at $t=1$ for $k=1$}
\centering
\begin{tabular}{cccc}
\toprule
\textbf{} & PINN & SA-PINN & Causal PINN \\
\midrule
\textbf{$u^*$} & 240.05908 & 137.150752 & $1.20048 \times 10^{-6}$ \\
\textbf{$\theta^*$} & 9.18397 & 6.56500 & $7.70788 \times 10^{-6}$ \\
\bottomrule
\end{tabular}
\label{tab4}
\end{table}

Table~\ref{tab4} presents the relative percentage error in predicting displacement and rotation for vanilla PINN, SA-PINN, and causal PINN. In the case of causal PINN, both quantities of interest, $u$, and $\theta$ exhibit errors in the magnitude of $10^{-6}$, demonstrating its accuracy. Conversely, vanilla PINN fails to adequately approximate the quantities of interest, as evidenced by a relative error percent of over $200{\%}$ for displacement and an error of approximately $9{\%}$ for rotation. Also, for the case of SA-PINN, the relative error percent for displacement is over $100{\%}$ and around $6{\%}$ for rotation. The results show that Causal PINN accurately predicts displacement and rotation for the Timoshenko beam model.

\subsection{Large space-time horizon}
In the following two experiments, we show the potential of transfer learning and predict the displacement and cross-sectional rotation in a larger domain. We utilize transfer learning for extrapolating. There are several benefits to knowing deflections on larger domains. Firstly, it provides a better understanding of the structural behavior of the beam under different loading conditions. By analyzing the deflection over larger lengths, engineers can assess the beam's overall stability and structural integrity, which is crucial for designing safe and reliable structures.

Secondly, calculating the deflection for extended domains allows for more accurate predictions of the behaviour of the beam in real-world scenarios. This information is valuable in various engineering applications such as building design, bridge construction, and aerospace engineering, where accurate deflection predictions are essential for ensuring the structural performance and safety of the final product.

Also, studying the deflection of the beam over a larger domain can help identify potential areas of weakness or excessive deformation. This knowledge enables engineers to make informed decisions about reinforcing certain sections or implementing design modifications to improve the overall performance and durability of the structure.

Furthermore, studying larger domains can optimize material usage and cost-effectiveness in construction projects. By accurately predicting deflection, engineers can optimize the size, shape, and materials used to construct beams, leading to more efficient designs and reduced material waste.

\begin{figure*} 
\centering
\includegraphics[width=0.48\columnwidth]{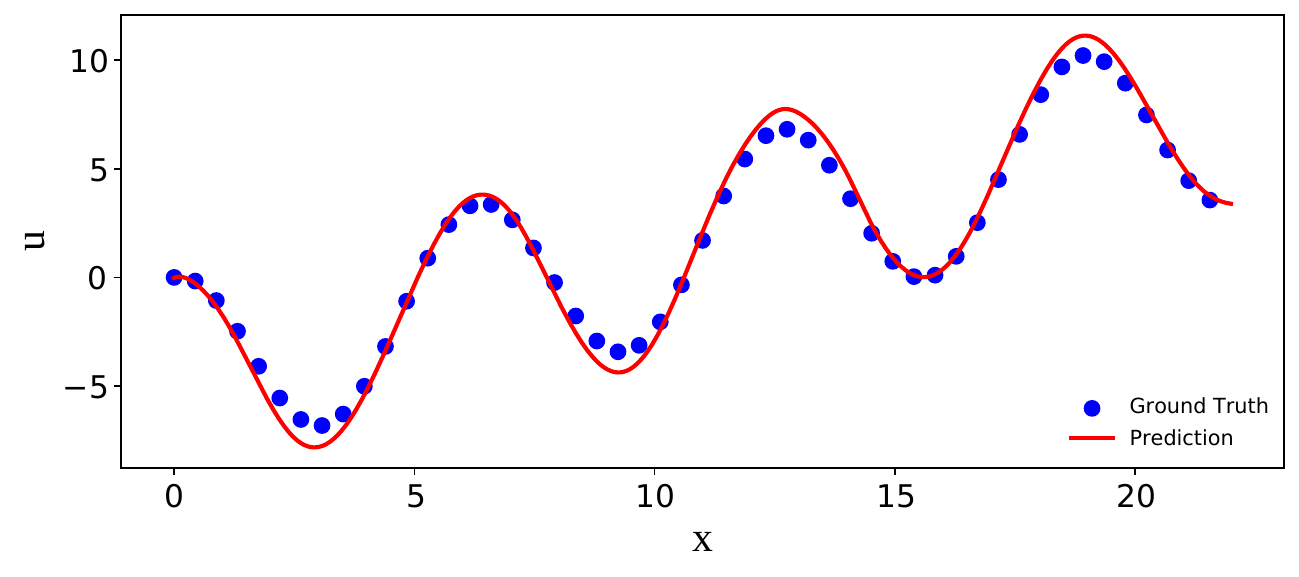}
\includegraphics[width=0.48\columnwidth]{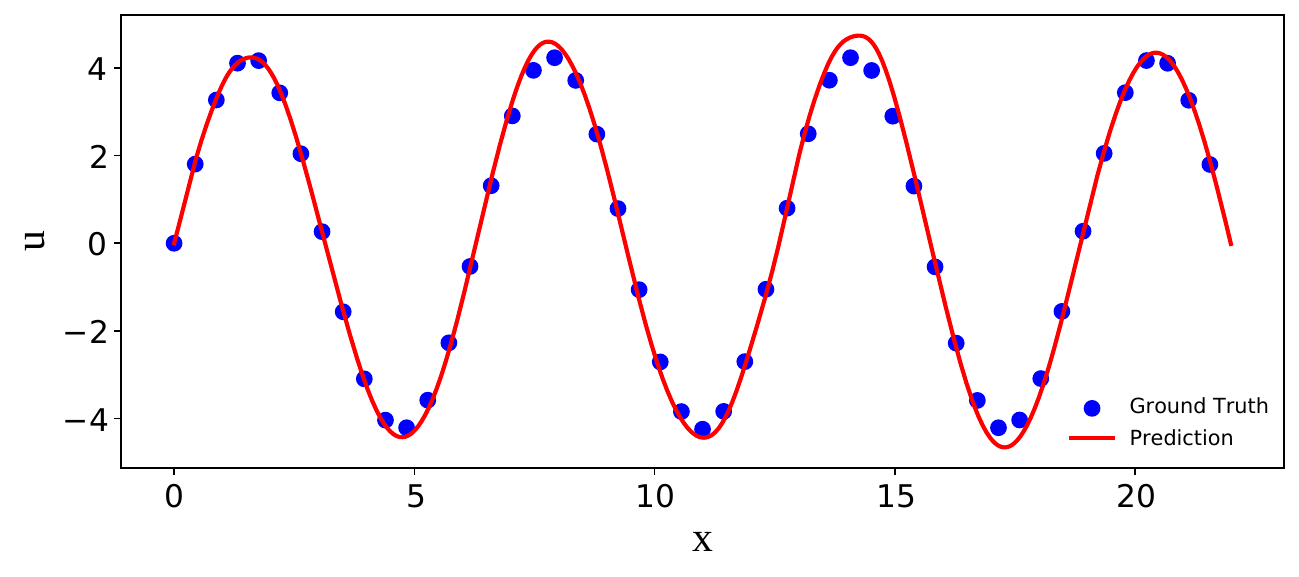}
\includegraphics[width=0.48\columnwidth]{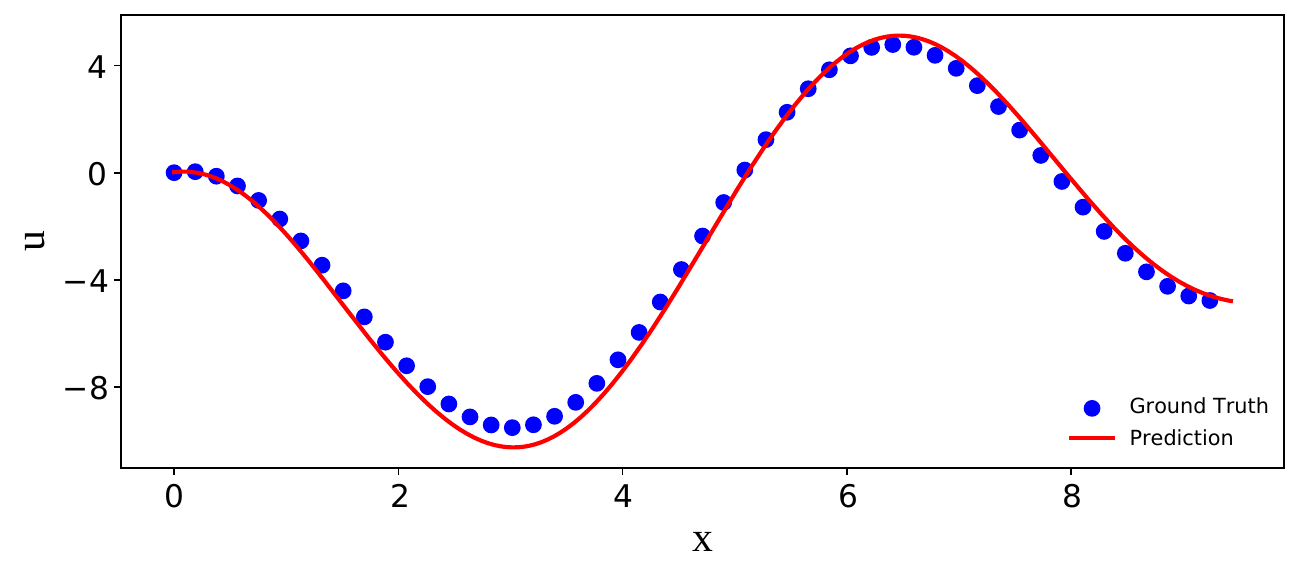}
\includegraphics[width=0.48\columnwidth]{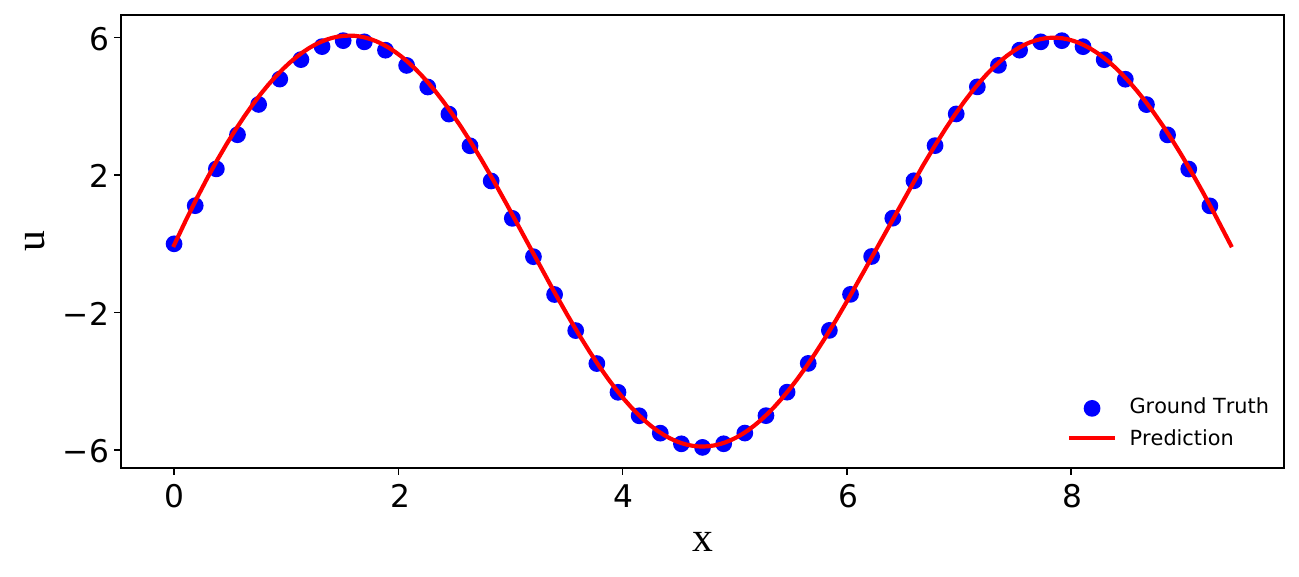}

\caption{Timoshenko beam on the Winkler foundation: \textbf{Top} Prediction for extended domain in space,  $x \in [0, 7 \pi]$ \textbf{Left:} Displacement ($u^{*}$); \textbf{Right:} Rotation ($\theta^{*}$). \textbf{Bottom} Prediction for extended domain in time for $t = 7$ \textbf{Left:} Displacement ($u^{*}$); \textbf{Right:} Rotation ($\theta^{*}$).}
  
\label{fig9}
\end{figure*}

\subsubsection{Extended spatial domain}
In this section, we consider the Timoshenko beam model in an extended domain in space. The spatial domain for the parent model is $x \in [0, 3\pi]$. Here, we utilize the parameters of the parent model and train the subsequent models for different spatial domains, in particular $x \in [0, 5\pi]$, $x \in [0, 6\pi]$, and $x \in [0, 7\pi]$. The aim is to observe the method's potential in a larger domain, indicating that the method generalizes well. The results obtained with and without transfer learning are presented in Table~\ref{tab5}, highlighting the superior accuracy achieved by the proposed method when utilizing parameters from the main model compared to training the model with Xavier initialization. Fig.~\ref{fig9} top row presents the proposed method's predictions of displacement and rotation, indicating that the model generalizes well across the spatial domain, inheriting the underlying structure and symmetry of the solution.

\begin{table}[htbp]
\centering
\caption{Timoshenko Beam: $\mathcal{R}$ for extension in the spatial domain with 3000 epochs}
\small
\begin{tabular}{ccccc}
\toprule
&\multicolumn{2}{c}{With TL}&\multicolumn{2}{c}{w/o TL} \\
\midrule
\textbf{$x$} & $u^*$ & $\theta^*$ & $u^*$ & $\theta^*$ \\
\midrule
$[0, 5 \pi]$ & $6.6 \times 10^{-5}$ & 0.00011 & 2.34306 & 3.51362 \\
$[0, 6 \pi]$ & 0.00653  & 0.00097 & 21.81964 & 30.67853 \\
$[0, 7 \pi]$ & 1.52043 & 0.61573 & 11.00256 & 8.90537 \\
\bottomrule
\end{tabular}
\label{tab5}
\end{table}

\subsubsection{Extended temporal domain}
We now extend our investigation to the temporal domain based on successfully generalizing the proposed method in the spatial domain. By employing the trained parameters obtained from the parent model, we train the same model with an extension in time, considering different temporal domains, $t \in [0, 4]$, $t \in [0, 6]$, and $t \in [0, 7]$. The relative error percentage for all cases of the extended temporal domains is presented in Table~\ref{tab6}. We observe that the proposed method accurately predicts displacement and rotation, while the approach without transfer learning fails to provide the same level of accuracy. Fig.~\ref{fig9} bottom row shows the predictions obtained by the proposed method for displacement and rotation in an extended temporal domain. The results show that utilizing transfer learning for extended domains in space and time provides accurate results, conserving the structure and symmetry of the solution.  

\section{Conclusions} \label{sec6}

This paper introduced a methodology for simulating the dynamics of beam models based on Euler-Bernoulli and Timoshenko's theories on the Winkler foundation. By incorporating transfer learning within a causality-respecting PINN framework, we addressed the need for re-training the network when there are modifications to the initial conditions or computational domain.

Numerical experiments demonstrated the effectiveness of the proposed approach. For the Euler-Bernoulli beam, we utilized the trained parameters from the parent model to simulate sub-cases with different initial conditions, including noisy ones. For the Timoshenko beam, we investigated its behavior in an extended spatial and temporal domain. These experiments showcased the generalization potential of the proposed method.

\begin{table}[htbp]
\centering
\caption{Timoshenko Beam: $\mathcal{R}$ for extension in the temporal domain with 3000 epochs}
\small
\begin{tabular}{ccccc}
\toprule
&\multicolumn{2}{c}{With TL}&\multicolumn{2}{c}{w/o TL} \\
\midrule
\textbf{$t$} & $u^*$ & $\theta^*$ & $u^*$ & $\theta^*$ \\
\midrule
$[0, 4]$ & 9.7e-6 & 2.4e-5 & 7.9e-5 & 0.00026 \\
$[0, 6]$ & 0.00111  & 0.00085 & 0.01627 & 0.12266 \\
$[0, 7]$ & 0.89122 & 0.05554 & 4.92954 & 2.50340 \\
\bottomrule
\end{tabular}
\label{tab6}
\end{table}

We also performed comparisons of the proposed method with SA-PINNs and vanilla PINNs. Results show that the causality-respecting PINN with transfer learning reduces computational costs and improves convergence. The results indicate that the method struggled to approximate the solutions accurately without transfer learning.

Overall, our findings highlight the efficacy of the proposed methodology in simulating beam dynamics under diverse engineering scenarios. By leveraging transfer learning and a causality-respecting PINN framework, we can reduce training requirements while achieving accurate results for various cases. This research opens up new possibilities for efficiently predicting the dynamics of structural elements, leading to advancements in structural engineering design, optimization, and control. 

Future research directions involve extending the methodology to other structural elements like systems of beams, strings and plates. An alternative research trajectory may involve training a family of PDE models and applying meta-learning techniques to derive a universal set of parameters applicable across diverse models. This unified parameter set could potentially be employed to test novel models, contributing to a generalized and efficient approach in the field. The codes will be made available upon publication.

\section*{CRediT authorship contribution statement}
\textbf{Taniya Kapoor:} Conceptualization, Methodology, Software, Validation, Formal analysis, Investigation, Writing - Original Draft, Writing - Review \& Editing, Visualization. \textbf{Hongrui Wang:} Writing - Review \& Editing, Supervision, Project administration. \textbf{Alfredo N\'u\~nez:} Writing - Review \& Editing, Supervision, Project administration. \textbf{Rolf Dollevoet:} Resources, Funding acquisition, Supervision, Project administration.

\section*{Declaration of competing interest}
The authors declare that they have no known competing financial interests or personal relationships that could have appeared to influence the work reported in this paper.

\section*{Acknowledgment}
The authors would like to acknowledge the support and computational resources provided by the DelftBlue Cluster contributing to this research.

\bibliographystyle{unsrt}  
\bibliography{references}  
\end{document}